\documentclass[10pt,twocolumn,letterpaper]{article}

\usepackage{cvpr}
\usepackage{times}
\usepackage{epsfig}
\usepackage{graphicx}
\usepackage{amsmath}
\usepackage{amssymb}
\usepackage{breqn}    
\usepackage{makecell} 
\usepackage{pgfplots} 

\usepackage[breaklinks=true,bookmarks=false]{hyperref}

\cvprfinalcopy 

\def\cvprPaperID{****} 

\begin{document}

\title{Learning Non-Metric Visual Similarity for Image Retrieval}

\author{Noa Garcia\\
Aston University\\
{garciadn@aston.ac.uk}
\and
George Vogiatzis\\
Aston University\\
{g.vogiatzis@aston.ac.uk}
}

\maketitle

\begin{abstract}
Measuring visual similarity between two or more instances within a data distribution is a fundamental task in image retrieval. Theoretically, non-metric distances are able to generate a more complex and accurate similarity model than metric distances, provided that the non-linear data distribution is precisely captured by the system. In this work, we explore neural networks models for learning a non-metric similarity function for instance search. We argue that non-metric similarity functions based on neural networks can build a better model of human visual perception than standard metric distances. As our proposed similarity function is differentiable, we explore a \textit{real} end-to-end trainable approach for image retrieval, i.e. we learn the weights from the input image pixels to the final similarity score. Experimental evaluation shows that non-metric similarity networks are able to learn visual similarities between images and improve performance on top of state-of-the-art image representations, boosting results in standard image retrieval datasets with respect standard metric distances.
\end{abstract}

\section{Introduction}

For humans, deciding whether two images are visually similar or not is, to some extent, a natural task. However, in computer vision, this is a challenging problem and algorithms do not always succeed in matching pictures that contain similar-looking elements. This is mainly because of the well-known \textit{semantic gap} problem, which refers to the difference or gap between low-level image pixels and high-level semantic concepts. Estimating visual similarity is a fundamental task that seeks to break this semantic gap by accurately evaluating how alike two or more pictures are. Visual similarity is crucial for many computer vision areas including image retrieval, image classification and object recognition, among others.

\begin{figure}
\centering
\setlength{\tabcolsep}{2pt}
\begin{tabular}{cc}
{\includegraphics[width = 0.245\textwidth]{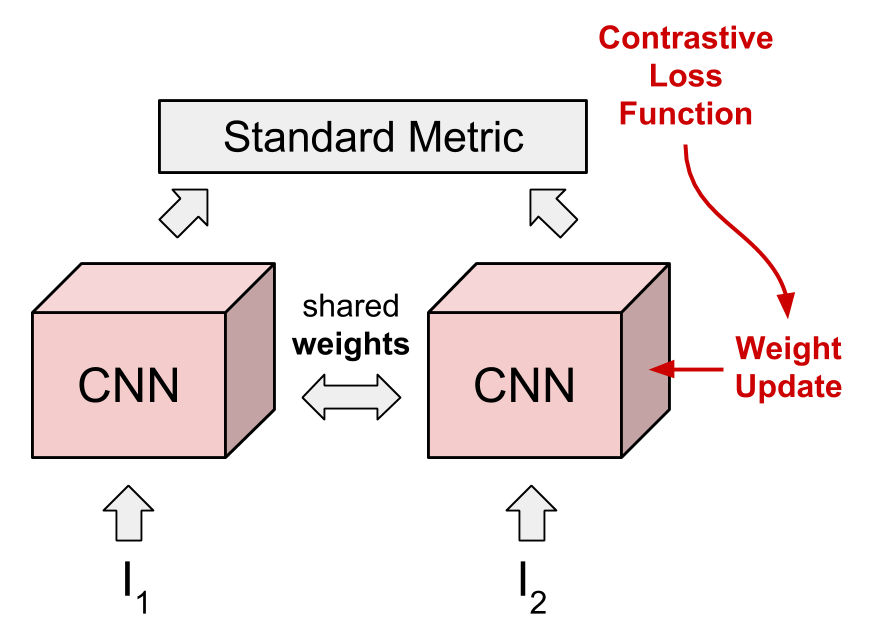}} &
{\includegraphics[width = 0.215\textwidth]{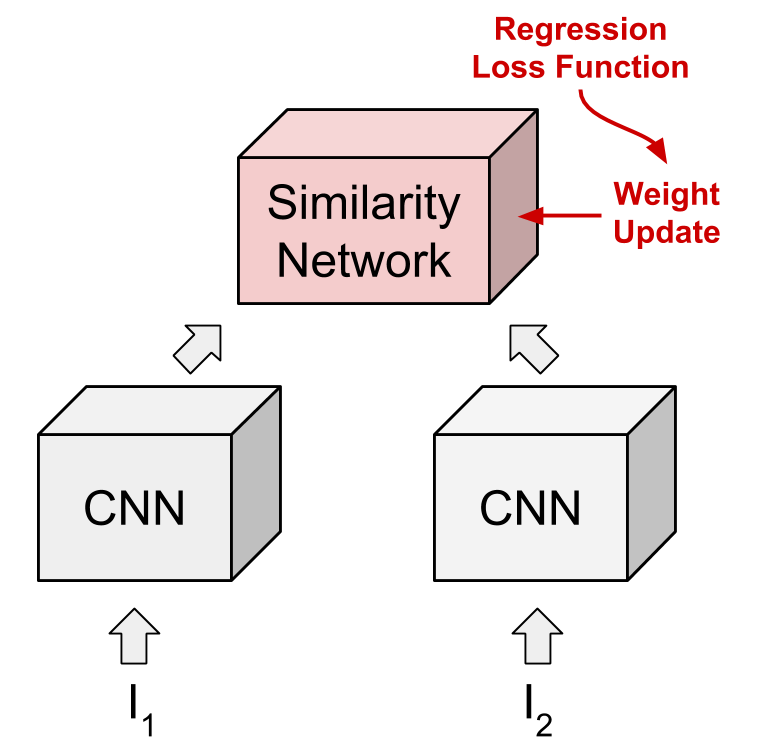}} \\
\end{tabular}
\caption{Similarity versus Siamese networks. Siamese networks (left) learn to map pixels into vector representations, whereas Similarity networks (right) learn a similarity function on top of the vector representations.}
\label{fig:siamese}
\end{figure}

Given a query image, content-based image retrieval systems rank pictures in a dataset according to how similar they are with respect to the input. This can be broken into two fundamental tasks: 1) computing meaningful image representations that capture the most salient visual information from pixels and 2) measuring accurate visual similarity between these image representations to rank images according to a similarity score.

In the last years, several methods to represent visual information from raw pixels in images have been proposed, first by designing handcrafted features such as SIFT \cite{lowe2004distinctive}, then by compacting these local features into a single global image descriptor using different techniques such as Fisher Vectors \cite{perronnin2010large} and more recently by extracting deep image representations from neural networks \cite{babenko2014neural}. However, once two images are described by feature vectors, visual similarity is commonly measured by computing a standard metric between them. Although regular distance metrics, such as Euclidean distance or cosine similarity, are fast and easy to implement, they do not take into account the possible interdependency within the dataset, which means that even if a strong nonlinear data dependency is occurring in the visual collection, they might not be able to capture it. This suggests that learning a similarity estimation directly from visual data can improve the performance on image retrieval tasks, provided that the likely nonlinearity dependencies within the dataset are precisely learned.

\begin{figure*}
\centering
\setlength{\tabcolsep}{2pt}
\begin{tabular}{cc}
\includegraphics[width = 0.5\textwidth]{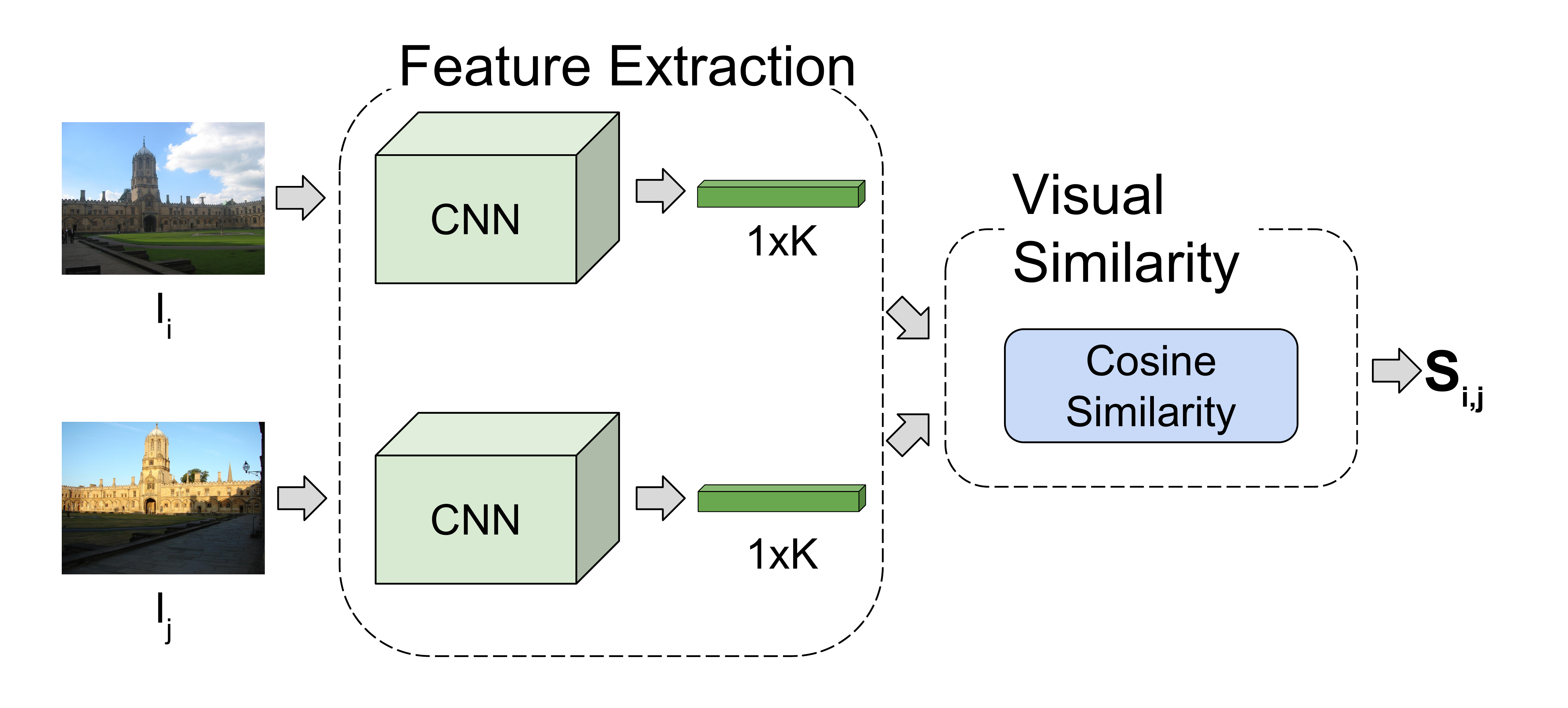} &
\includegraphics[width = 0.5\textwidth]{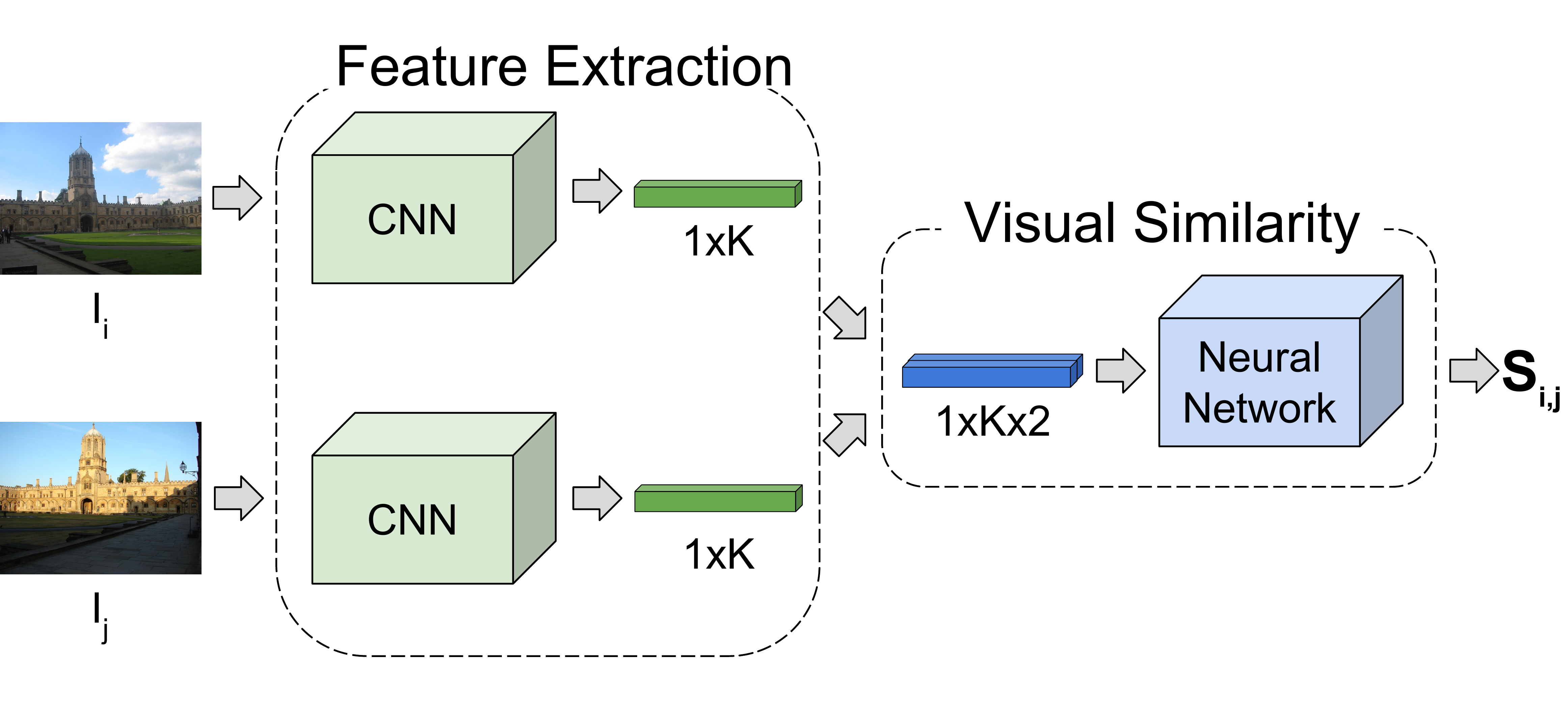} \\
\small (a) & \small (b) \\
\end{tabular}
\caption{Retrieval system based on metric distances versus our proposed model. (a) Standard systems use a metric distance function to estimate the visual similarity between a pair of feature vectors obtained from a feature extraction model. (b) Our proposed model estimates the similarity score, $S_{i,j}$, from a pair of visual vectors by using a non-metric similarity network.}
\label{fig:overview}
\end{figure*}

In this work, we propose a model to learn a non-metric visual similarity function on top of image representations for pushing accuracy in image retrieval tasks. This idea is shown in Figure \ref{fig:overview}. As in standard image retrieval systems, we extract $K$-dimensional visual vectors from images by using a convolutional neural network (CNN) architecture. Then, a \textit{visual similarity neural network} is used to estimate the similarity score between a pair of images. Note that in standard systems this score is usually computed with a metric distance. We design a supervised regression learning protocol so that different similarity degrees between images are precisely captured. Then, we directly apply the output of the model as a similarity estimation to rank images accordingly. In this way, the similarity network can be seen as a replacement of the standard metric distance computation, being able to mathematically fit visual human perception better than standard metrics and to improve results on top of them. The proposed similarity network is end-to-end differentiable, which allows us to build an architecture for \textit{real} end-to-end training: from the input image pixels to the final similarity score. Experimental evaluation shows that performance on standard image retrieval datasets is boosted when the similarity function is directly learnt from the visual data instead of using a rigid metric distance. 

Many techniques have been envisaged to boost image retrieval performance in the past, such as query expansion and re-ranking \cite{arandjelovic2012three}, network fine-tunning \cite{babenko2014neural,gordo2017end} or feature fusion \cite{dong2018late,dong2018few}. However, these techniques are not competitors of our method but optional add-ons, as we argue that the methodology proposed in this work can be applied along with all of them in the same way as they are being applied on systems based on metric distances. Moreover, training a similarity network as we propose is computationally simpler than fine-tuning the whole feature representation network (i.e. network fine-tunning), extracting multiple features using different networks (i.e. feature fusion) or computing multiple queries per image (i.e. query expansion and re-ranking).

In summary, the main contributions of this work are:
\begin{enumerate}
\setlength\itemsep{0em}
\item We present a neural network architecture to model visual similarities, which introduces a new and simple method to boost performance in image retrieval by only training the last stage of the system.
\item We propose a novel regression loss function specifically designed for improving similarity scores on top of standard metrics in image retrieval tasks.
\item We design a \textit{real} end-to-end system for content-based image retrieval that can be trained from the input image pixels to the final similarity score.
\item We empirically show the efficacy of our method in standard image retrieval datasets. Via our ablation study, we show that the proposed system can successfully compute visual similarities on top of different standard retrieval features, outperforming cosine similarity and metric learning in most of the datasets.
\end{enumerate}


\section{Related Work}
\label{sec:relatedwork}

\paragraph{Content-Based Image Retrieval} Content-based image retrieval searches for images by considering their visual content. Given a query image, pictures in a collection are ranked according to their visual similarity with respect to the query. Early methods represent the visual content of images by a set of hand-crafted features, such as SIFT \cite{lowe2004distinctive}. As a single image may contain hundreds of these features, aggregation techniques like bag-of-words (BOW) \cite{sivic2003video}, Fisher Vectors \cite{perronnin2010large} or VLAD \cite{jegou2012aggregating} encode local descriptors into a compact vector, thereby improving computational efficiency and scalability. More recently, because of the latest advancements on deep learning, features obtained from convolutional neural networks (CNN) have rapidly become the new state-of-the-art in image retrieval.

\paragraph{Deep Learning for Image Retrieval} Deep image retrieval extracts activations from CNNs as image representations. At first, some methods \cite{babenko2014neural, razavian2014cnn, wan2014deep, liu2015deepindex} proposed to use representations from one of the last fully connected layers of networks pre-trained on the classification ImageNet dataset \cite{ILSVRC15}. When deeper networks such as GoogLeNet \cite{szegedy2015going} and VGG \cite{simonyan2014very} appeared, some authors \cite{babenko2015aggregating, yue2015exploiting,razavian2014cnn, xie2015image} showed that mid-layer representations obtained from the convolutional layers performed better in the retrieval task. Since then, there have been several attempts to aggregate these high-dimensional convolutional representations into a compact vector. For example, \cite{gong2014multi, yue2015exploiting} compacted deep features by using VLAD, \cite{mohedano2016bags} encoded the neural codes into an histogram of words, \cite{babenko2015aggregating, kalantidis2016cross} applied sum-pooling to obtain a compact representation and \cite{razavian2016visual, tolias2016particular} aggregated deep features by max-pooling them into a new vector. A different approach is to train the network to directly learn compact binary codes end-to-end \cite{erin2015deep, lin2015deep}. Some authors have shown that fine-tunning the networks with similar data to the target task increases the performance significantly \cite{babenko2014neural, gordo2016deep, radenovic2016cnn, salvador2016faster,gordo2017end}. Finally, recent work has shown that adding attention models to select meaningful features can be also beneficial for image retrieval \cite{jimenez2017class, Noh2017Large}.

All of these methods are focused on finding high quality features to represent visual content efficiently and visual similarity is computed by simply applying a standard metric distance. General metrics, such as Euclidean distance or cosine similarity, however, might be failing to consider the inner data structure of these visual representations. Learning a similarity function directly from data may help to capture the human perception of visual similarity in a better way.

\paragraph{Similarity Learning} 
Some of the most popular similarity learning work, such as OASIS \cite{chechik2010large} and MLR \cite{mcfee2010metric}, are based on linear metric learning by optimizing the weights of a linear transformation matrix. For example, Yang et al. \cite{yang2012multimedia} proposed a framework for ranking elements in retrieval tasks by solving a linear optimization problem. Although linear methods are easier to optimize and less prone to overfitting, nonlinear algorithms are expected to achieve higher accuracy modeling the possible nonlinearities of data. 

Nonlinear similarity learning based on deep learning has been recently applied to many different visual contexts. In low-level image matching, CNNs have been trained to match pairs of patches for stereo matching \cite{zagoruyko2015learning, luo2016efficient} and optical flow \cite{fischer2015flownet, thewlis16fully-trainable}. In high-level image matching, deep learning techniques have been proposed to learn low-dimensional embedding spaces in face verification \cite{chopra2005learning}, retrieval \cite{wu2013online,wang2014learning}, classification \cite{hoffer2015deep, qian2015fine, song2016deep} and product search \cite{bell2015learning}, either by using siamese \cite{chopra2005learning} or triplet \cite{wang2014learning} architectures. More recently, deep similarity learning has also been applied to fabric image retrieval \cite{deng2018learning} by using triplets of samples to ensure that similar features are mapped closer than non-similar features.

In general, these methods rely on learning a mapping from image pixels to a low dimensional target space to compute the final similarity decision by using a standard metric. They are designed to find the best projection in which a linear distance can be successfully applied. Instead of projecting the visual data into some linear space, that may or may not exist, our approach seeks to learn the non-metric visual similarity score itself. Similarly, \cite{li2014deepreid} and \cite{han2015matchnet} used a CNN to decide whether or not two input images are a match, applied to pedestrian re-identification and patch matching, respectively. In these methods, the networks are trained as a binary classification problem (i.e. same or different pedestrian/patch), whereas in an image retrieval ranking problem, a regression score is required. Inspired by the results of \cite{wan2014deep}, which showed that combining deep features with similarity learning techniques can be very beneficial for the performance of image retrieval systems, we propose to train a deep learning algorithm to learn non-metric similarities for image retrieval and improve results in top of high quality image representation methods.

Neural networks have been previously proposed to model relationships between objects in different domains, such as classification in few-shot learning \cite{vinyals2016matching,santoro2017simple} or visual question answering \cite{sung2018learning}. The main difference between these methods and this work lies in the optimization problem to be solved: whereas \cite{vinyals2016matching,santoro2017simple, sung2018learning} aim to learn if a certain relation between a pair of images is occurring by optimizing a classification loss function (e.g. whether a query image is similar to a samples from a known class or not), we introduce a novel loss function specifically designed to solve ranking problems, which improves similarity scores on top of a standard metric by returning a regression value (i.e. how similar the two images are).


\section{Methodology}
\label{sec:method}
In this section, we present our proposed method to learn a non-metric visual similarity function from the visual data distribution for image retrieval.

\begin{figure*}
\centering
\includegraphics[width = 0.8\textwidth]{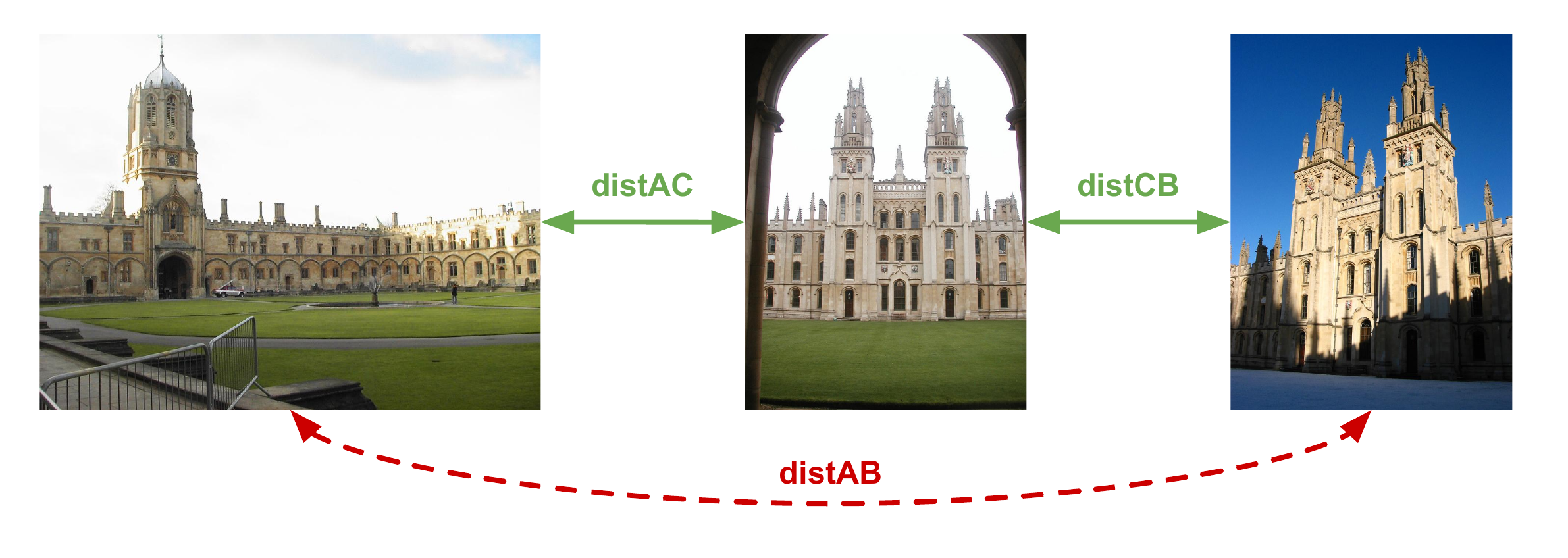}
\caption{Triangle inequality does not necessarily fit human visual perception, as distAB is expected to be bigger than distAC plus distCB.}
\label{fig:triangle}
\end{figure*}

\subsection{Visual Similarity}
Visual similarity measures how alike two images are. Formally, given a pair of images $I_i$ and $I_j$ in a collection of images $\xi$, we define $s_{i,j}$ as their similarity score. The higher $s_{i,j}$ is, the more similar $I_i$ and $I_j$ are. To compute $s_{i,j}$, images are represented by $K$-dimensional image representations, which are obtained by mapping image pixels into the feature space $\mathbb{R}^K$, as $x_k = f(I_k,w_f)$ with $I_k \in \xi$, where $f(\cdot)$ is a non-linear image representation function and $w_f$ its parameters. We propose to learn a visual similarity function, $g(\cdot)$, that maps a pair of image representations $x_i$ and $x_j$ into a visual score as:%
\begin{multline}
s_{i,j} = g(f(I_i,w_f),f(I_j,w_f),w_g) \\
\quad s.t. \quad s_{i,j} > s_{i,k} \rightarrow I_i, I_j \text{ more} \\ 
\text{similar than } I_i, I_k
\label{eq:simscore}
\end{multline}%
with $I_i, I_j, I_k \in \xi$ and $w_g$ being the trainable parameters of the similarity function. 

Visual similarity functions are commonly based on metric distance functions such as $g(x_i,x_j) = \frac{x_i\cdot x_j}{\|x_i \|\|x_j \|}$ or $g(x_i,x_j) = \|x_i - x_j \| $, i.e. cosine similarity and Euclidean distance, respectively. Metric distance functions, $d(\cdot)$, perform mathematical comparisons between pairs of objects in a collection $\Pi$, by satisfying the following axioms:  
\begin{enumerate}
\item $d(a,b) \geq 0$ (non-negativity)
\item $d(a,b) = 0 \leftrightarrow a = b$ (identity)
\item $d(a,b) = d(b,a)$ (symmetry)
\item $d(a,b) \leq d(a,c) + d(c,b)$ (triangle inequality)
\end{enumerate} %
with  $\forall a,b,c \in \Pi$. 

However, metric axioms are not always the best method to represent visual human perception \cite{gavet2014dissimilarity, tan2006learning, tversky1982similarity}. For example, non-negative and identity axioms are not required in visual perception as long as relative similarity distances are maintained. Symmetry axiom is not always true, as human similarity may be influenced by the order of the objects being compared. Finally, triangle inequality does not correspond to visual human perception either. It can be easily seen when considering the images of a person, a horse and a centaur: although a centaur might be visually similar to both a person and a horse, the person and the horse are not similar to each other. A visual example applied to image retrieval can be seen in Figure \ref{fig:triangle}. 

For a better mathematical representation of the visual human perception, we propose to learn a non-metric visual similarity function without requiring to satisfy the rigid metric axioms.

\subsection{Similarity Network}
To fit visual human perception better than metric distance functions, we propose to learn a similarity function from the visual data using neural networks. This similarity network is composed of a set of fully connected layers, each one of them, except by the last one, followed by a ReLU \cite{krizhevsky2012imagenet} non-linearity.

The input of the network is a concatenated pair of image representations vectors, $x_i$ and $x_j$, which can be obtained using any standard technique, such as \cite{babenko2014neural, tolias2016particular} or any other. The output is a similarity score, $s_{i,j}$. In that way, the similarity network learns the similarity function, $g(\cdot)$, from the image representation vectors and replaces the metric distance function used in standard systems. 

At this point, we would like to note that the proposed similarity network is conceptually different to the siamese architecture in \cite{chopra2005learning}, as shown in Figure \ref{fig:siamese}. Siamese networks use pairs of images to learn the feature extraction function, $f(\cdot)$, which maps image pixels images into vector representations. Then, similarity is still computed with a metric distance function, such as cosine similarity or Euclidean distance. In contrast, our approach learns the function $g(\cdot)$ on top of the image representations, replacing the standard metric distance computation. 

\begin{figure*}
\centering
\setlength{\tabcolsep}{3pt}
\begin{tabular}{c c @{\hspace{20pt}} c c}
{\includegraphics[width = 0.23\textwidth]{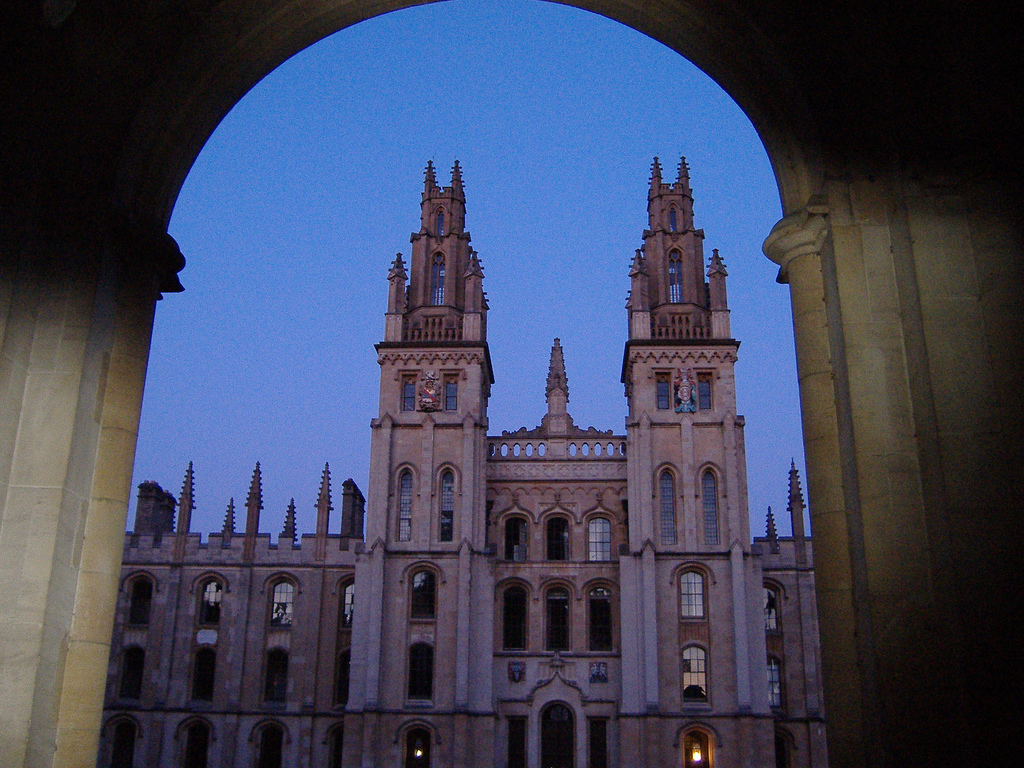}} &
{\includegraphics[width = 0.23\textwidth]{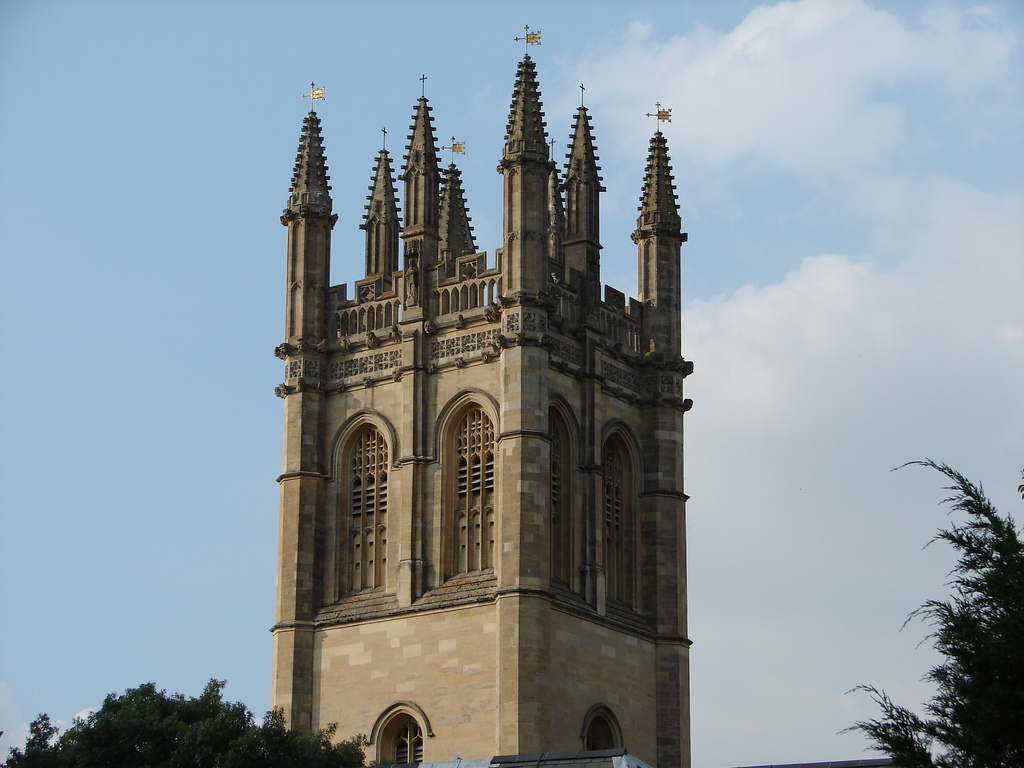}} & 
{\includegraphics[width = 0.23\textwidth]{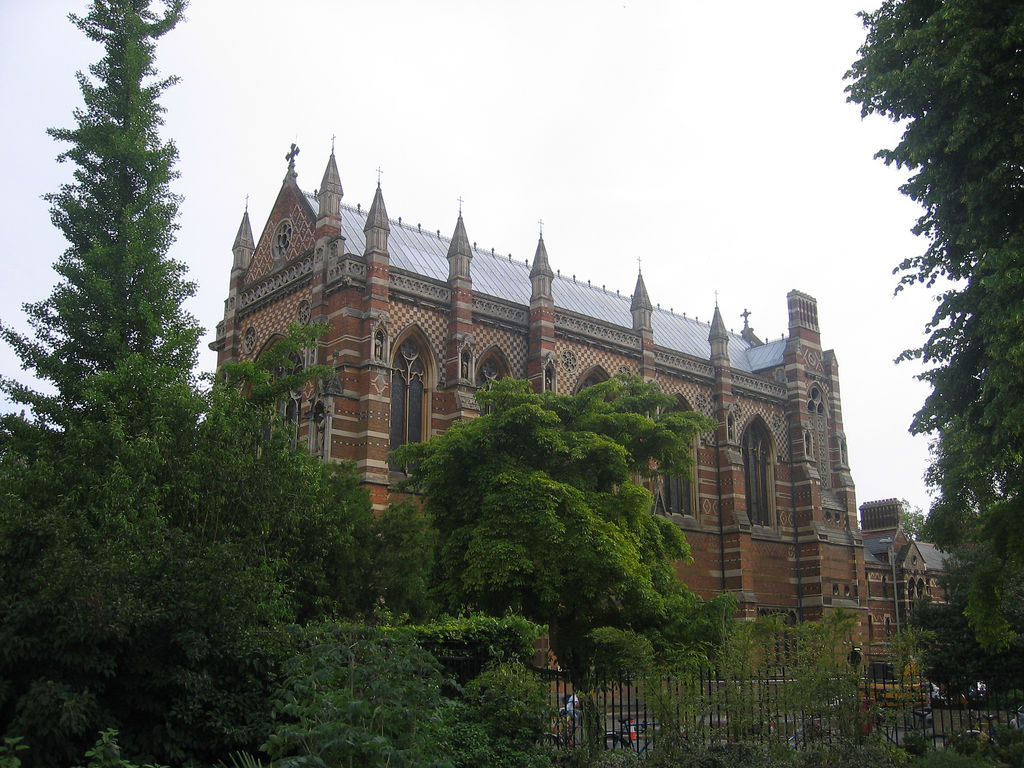}} &
{\includegraphics[width = 0.23\textwidth]{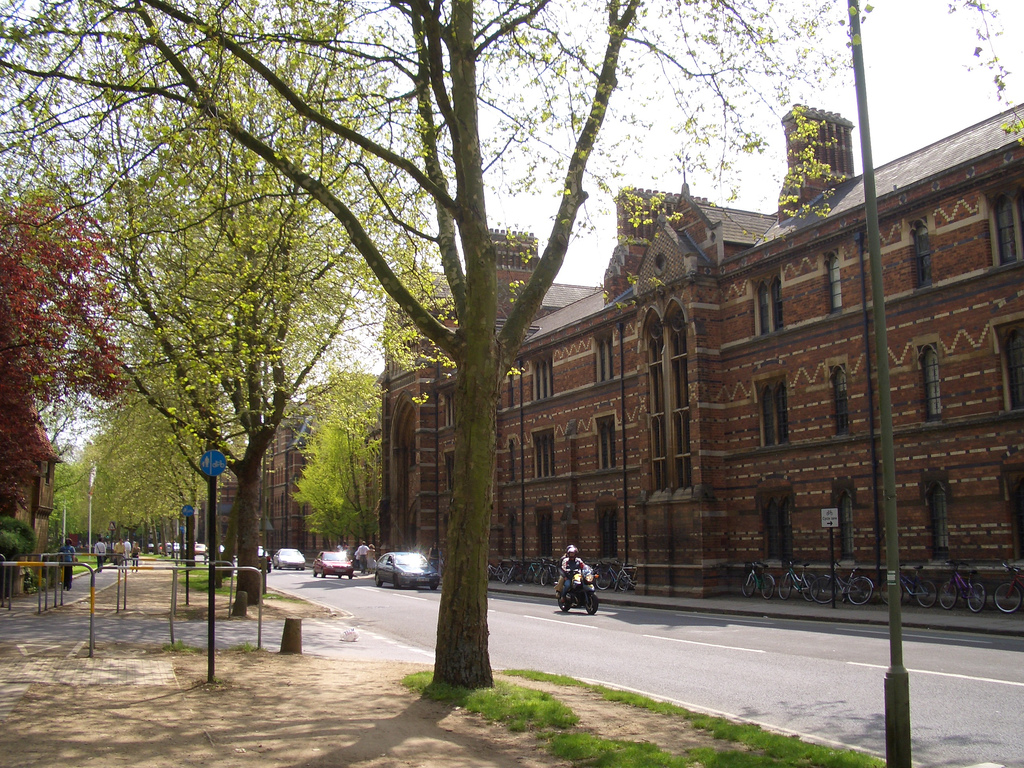}}\\[5pt]
\multicolumn{2}{c}{Dissimilar images} & \multicolumn{2}{c}{Similar images} \\
\end{tabular}
\vspace{5pt}
\caption{Example of difficult pairs. Dissimilar images with lower score than the metric distance (left) and similar images with higher score than the metric distance (right).}
\label{fig:difficult}
\end{figure*}

\subsection{Similarity Training}
\label{sec:stages}
We design the training of the similarity network as a supervised regression task. However, as providing similarity labels for every possible pair of training images is infeasible, we propose a training procedure in which the visual similarity is learned progressively using standard image classification annotations. The model is trained to discriminate whether two images, $I_i,I_j$, are similar or dissimilar. Then, a similarity score, $s_{i,j}$, is assigned accordingly by improving a standard similarity function, $sim(\cdot)$. To optimize the weights, $w_g$, of the similarity function $g(\cdot)$ from Equation \ref{eq:simscore}, the following regression loss function is computed between each training pair of image representations, $x_i,x_j$: %
\begin{dmath}
\mathcal{L}(I_i,I_j) = |s_{i,j} - \ell_{i,j} (sim(x_i,x_j) + \Delta) - (1-\ell_{i,j})(sim(x_i,x_j) - \Delta)| 
\label{eq:loss}
\end{dmath} where $\Delta$ is a margin parameter and $\ell_{i,j}$ is defined as:

\begin{equation}
\ell_{i,j}=
\begin{cases}
    1   & \text{if } I_i \text{ and } I_j \text{ are similar}\\
    0  	& \text{otherwise}\\
\end{cases}  
\end{equation}%

In other words, the similarity network learns to increase the similarity score when two matching images are given and to decrease it when a pair of images is not a match. Similarity between pairs might be decided using different techniques, such as image classes, score based on local features or manual labeling, among others \cite{bhagat2018image}. Without loss of generality, we consider two images as similar when they belong to the same annotated class and as dissimilar when they belong to different classes.

Choosing appropriate examples when using pairs or triplets of samples in the training process is crucial for a successful training \cite{gordo2016deep,radenovic2016cnn,movshovitz2017no}. This is because if the network is only trained by using \textit{easy pairs} (e.g. a car and a dog), it will not be able to discriminate between \textit{difficult pairs} (e.g. a car and a van). We design our training protocol by emphasizing the training of difficult examples. First, we randomly select an even number of similar and dissimilar pairs of training samples and train the similarity network until convergence. We then choose a new random set of images and compute the similarity score between all possible pairs by using the converged network. Pairs in which the network output is worse than the metric distance function measure are selected as difficult pairs for retraining, where a worse score means a score that is lower in the case of a match and higher in the case of a non-match. Finally, the difficult pairs are added to the training process and the network is trained until convergence one more time. Examples of difficult image pairs are shown in Figure \ref{fig:difficult}.

\section{Experimental Evaluation}
\label{sec:experiments}
We present the experimental evaluation we perform to validate the proposed non-metric similarity network.

\subsection{Image Retrieval Datasets}
\label{sec:datasets}
Our approach is evaluated on three standard image retrieval datasets: \textsc{Oxford5k} \cite{Philbin07}, \textsc{Paris6k} \cite{Philbin08} and \textsc{Land5k}, a validation subset of \textsc{Landmarks} \cite{babenko2014neural} dataset. \textsc{Oxford5k} consists on 5,062 images of 11 different Oxford landmarks and 55 query images. \textsc{Paris6k} contains 6,412 images of 11 different Paris landmarks and 55 queries. \textsc{Land5k} consists of 4,915 images from 529 classes with a random selection of 45 images to be used as queries. For experiments on larger datasets, we also use the standard large-scale versions \textsc{Oxford105k} and \textsc{Paris106k}, by including 100,000 distractor images \cite{Philbin07}. In both \textsc{Oxford5k} and \textsc{Paris6k} collections query images are cropped according to the region of interest and evaluation is performed by computing the mean Average Precision (mAP). For \textsc{Land5k} results are also reported as mAP, considering an image to be relevant to a query when they both belong to the same class.

For training, we use the cleaned version of the \textsc{Landmarks} \cite{babenko2014neural} dataset from \cite{gordo2016deep}. Due to broken URLs, we could only download 33,119 images for training and 4,915 for validation. To ensure visual similarity is learnt from relevant data, we create two more training sets, named \textsc{Landmarks-extra500} and \textsc{Landmarks-extra}, by randomly adding about 500 and 2000 images from \textsc{Oxford5k} and \textsc{Paris6k} classes to \textsc{Landmarks}, respectively. Query images are not added in any case and they remain unseen by the system.

\begin{table*}
\centering
\begin{tabular}[t]{ c l  c  c  c }
	\Xhline{2\arrayrulewidth}
    \normalsize Config	 & \normalsize Architecture & \normalsize Params & \normalsize MSE & \normalsize $\rho$ \\ \hline
    A &  FC-1024, FC-1024, FC-1 & 2.1M	& 0.00035 & 0.909   \\ 
    B &  FC-4096, FC-4096, FC-1 & 21M	& 0.00019 & 0.965  \\ 
    C &  FC-8192, FC-8192, FC-1 & 76M 	& 0.00012 & 0.974   \\ 
    D &  FC-4096, FC-4096, FC-4096, FC-1 & 38M	& 0.00019 & 0.964  \\ 
    \Xhline{2\arrayrulewidth}
\end{tabular}
\vspace{5pt}
\caption{Architecture Discussion. Fully connected layers are denoted as (FC-\{filters\}).}
\label{tab:Configurations}
\end{table*}

\subsection{Experimental Details}
\label{sec:details}

\paragraph{Image Representation} Unless otherwise stated, we use RMAC \cite{tolias2016particular} as image representation method. VGG16 network is used off-the-shelf without any retraining or fine-tunning. Images are re-scaled up to 1024 pixels, keeping their original aspect ratio. RMAC features are very sensitive to the PCA matrices used for normalization. For consistency, we use the PCA whitening matrices trained on \textsc{Paris5k} on all the datasets, instead of using different matrices in each testing collection. This leads to slightly worse results than the ones provided in the original paper.

\paragraph{Similarity Training} We use cosine similarity, $sim(x_i,x_j) = \frac{x_i\cdot x_j}{\|x_i \|\|x_j \|}$, as the similarity function in Equation \ref{eq:loss}. For a faster convergence, we warm-up the weights of the similarity network by training it with random generated pairs of vectors and $\Delta = 0$. In this way, the network first learns to imitate the cosine similarity. Visual similarity is then trained using almost a million of image pairs. We experiment with several values of the margin parameter $\Delta$, ranging from 0.2 to 0.8. The network is optimized using backpropagation and stochastic gradient descent with a learning rate of 0.001, a batch size of 100, a weight decay of 0.0005 and a momentum of 0.9. 

\paragraph{Computational cost} Standard metric functions are relatively fast and computationally cheap. Our visual similarity network involves the use of millions of parameters that inevitable increase the computational cost. However, it is still feasible to compute the similarity score in a reasonable amount of time. In our experiments, training time is about 5 hours in a GeForce GTX 1080 GPU without weight warm-up and testing time for a pair of images is 1.25 ms on average (0.35 ms when using cosine similarity in CPU).

\subsection{Architecture Discussion}
We first experiment with four different architectures (Table \ref{tab:Configurations}) and compare the performance of each configuration during the network warm-up ($\Delta=0$), by using 22.5 million and 7.5 million pairs of randomly generated vectors for training and  validation, respectively. During the training warm-up, the network is intended to imitate the cosine similarity. We evaluate each architecture by computing the mean squared error, MSE, and the correlation coefficient, $\rho$, between the network output and the cosine similarity. Configuration C, which is the network with larger number of parameters, achieves the best MSE and $\rho$ results. Considering a trade-off between performance and number of parameters of each architecture, we keep configuration B as our default architecture for the rest of the experiments.

\begin{table*}
\centering
\vspace{5pt}
\begin{tabular}{ r  c  c  c  c  c  c  c  c  c} 
\Xhline{2\arrayrulewidth}
    & \multicolumn{3}{c}{\textsc{Landmarks}}
    & \multicolumn{3}{c}{\textsc{Landmarks-extra500}}
    & \multicolumn{3}{c}{\textsc{Landmarks-extra}} \\  \cline{2-10} 
        & \small Ox5k & \small Pa6k & \small La5k & \small \textsc{Ox5k} & \small \textsc{Pa6k} & \small \textsc{La5k} & \small \textsc{Ox5k} & \small \textsc{Pa6k} & \small \textsc{La5k} \\ \hline         
        \small Cosine	& \textbf{0.665} & 0.638 & 0.564 & 0.665 & 0.638 & 0.564  & 0.665 & 0.638 & 0.564 \\
    \small OASIS & 0.514 & 0.385 & 0.578 & 0.570  & 0.651  & 0.589 & 0.619 & 0.853 & 0.579 \\
    \small Linear \scriptsize{(0.2)}	& 0.598 & \textbf{0.660} & 0.508 & 0.611 & 0.632 & 0.514  & 0.602  & 0.581  & 0.502 \\
    \small SimNet \scriptsize{(0.2)} & 0.658	& 0.460 & 0.669 & 0.717 & 0.654  & 0.671 & 0.718 & 0.757 & 0.668 \\ 
    \small SimNet* \scriptsize{(0.2)}	& 0.655 & 0.503	& 0.697 & \textbf{0.719} & 0.677  & 0.693 & 0.786 & 0.860 & 0.662 \\
    \small SimNet* \scriptsize{(0.4)} & 0.637 & 0.504 & 0.737 & 0.703 & 0.701  & 0.745 & 0.794 & 0.878 & 0.706 \\ 
    \small SimNet* \scriptsize{(0.6)} & 0.613 & 0.514 & 0.776 & 0.703 & \textbf{0.716}  & 0.776 & 0.789 & 0.885 & 0.735  \\ 
    \small SimNet* \scriptsize{(0.8)} & 0.600 & 0.511 & \textbf{0.783} & 0.685 & 0.710  & \textbf{0.803} & \textbf{0.808} & \textbf{0.891} & \textbf{0.758} \\
    \Xhline{2\arrayrulewidth}
        \end{tabular}
\vspace{5pt}
\caption{Similarity Evaluation. mAP when using different similarity functions. $\Delta$ value is set in brackets.}
\label{tab:tablefull}
\end{table*}

\subsection{Similarity Evaluation}
\label{sec:evaluation}

We then study the benefits of using a non-metric similarity network for image retrieval by comparing it to several similarity methods. RMAC \cite{tolias2016particular} is used as image representation method in all the experiments. Similarity functions under evaluation are:

\paragraph{Cosine} The similarity between a pair of vectors is computed with the cosine similarity: $cosine(x_i, x_j) = \frac{x_i\cdot x_j}{\|x_i \|\|x_j \|}$. No training is required.

\paragraph{OASIS} Well-established OASIS algorithm \cite{chechik2010large} is used to learn a linear function to map a pair of vectors into a similarity score. The training of the matrix transformation is performed in a supervised way by providing the class of each image.

\paragraph{Linear} We learn an affine transformation matrix to map a pair of vectors into a similarity score by optimizing Equation \ref{eq:loss} in a supervised training. Classes of images are provided during training. The margin $\Delta$ is set to 0.2.

\paragraph{SimNet, SimNet*} The similarity function is learnt with our proposed similarity network by optimizing Equation \ref{eq:loss} with (SimNet*) or without (SimNet) difficult pairs refinement. Classes of images are provided during training and different margin $\Delta$ are tested, ranging from 0.2 to 0.8.

\pgfplotstableread[row sep=\\,col sep=&]{
    feature & Cosine & SimNet & SimNetD \\
    VGG16       & 0.480  & 0.516 & 0.554  \\
    RES50       & 0.176  & 0.243 & 0.471  \\
    MAC         & 0.539 & 0.683  & 0.750  \\
    RMAC        & 0.638 & 0.757 &  0.891 \\
    [26]        & 0.838 & 0.872 & 0.882 \\
    }\mydata
    
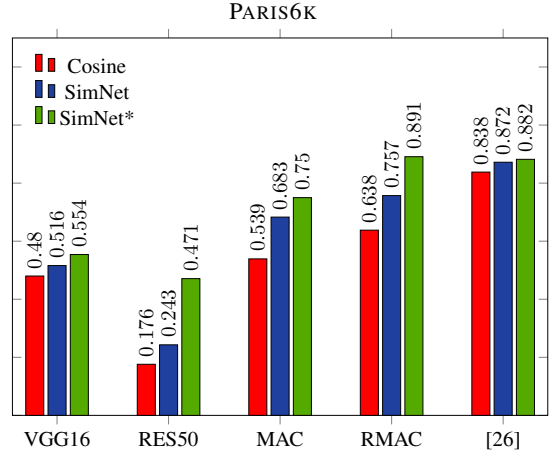
\begin{figure}
\centering
\begin{tikzpicture}[scale=0.8]
    \begin{axis}[
            legend style={draw=none, fill=none},
            ybar,
            bar width=.3cm,
            width=0.6\textwidth,
            height=.45\textwidth,
            legend pos=north west,
            symbolic x coords={VGG16, RES50, MAC, RMAC, [26]},
            xtick=data,
            nodes near coords,
            nodes near coords align={vertical},
            ymin=0,ymax=1.3,
            every node near coord/.append style={rotate=90, anchor=west},
            nodes near coords style={/pgf/number format/.cd,precision=3},
            yticklabels={,,}
        ]
        \addplot[fill=red,] table[x=feature,y=Cosine]{\mydata};
        \addplot[fill={rgb:red,1;green,2;blue,5},] table[x=feature,y=SimNet]{\mydata};
        \addplot[fill={rgb:red,1;green,2;blue,0},] table[x=feature,y=SimNetD]{\mydata};
        \legend{Cosine, SimNet, SimNet*}
    \end{axis}
    \node[above,font=\small] at (current bounding box.north) {\textsc{Paris6k}};
\end{tikzpicture}
\caption{mAP for different visual similarity techniques on top of different feature extraction methods.}
\label{fig:features}
\end{figure}

\begin{figure*}
\hspace*{-30pt}
\centering
\setlength{\tabcolsep}{10pt}
\begin{tabular}{cc}
\begin{tikzpicture}[scale=0.7]
\large
\begin{axis}[
    ylabel={mAP},
    xlabel={num. samples},
    xmin=0, 
    xmax=250,
    ymin=0.625, 
    ymax=0.725,
    xtick={0,50,100,150,200, 250},
    ytick={0.625, 0.65, 0.675, 0.70, 0.725},
    legend pos=north west,
    ymajorgrids=true,
    grid style=dashed,
    legend style={draw=none, fill=none},
]
 
\addplot[
    color={rgb:red,1;green,2;blue,5},
    line width=0.5mm
    ]
    coordinates {
    (0,0.658)(10, 0.659)(50, 0.659)(100, 0.674)(250,0.717)
    };
    
\addplot[
    color=red,
    dashed,
    line width=0.5mm
    ]
    coordinates {
    (0,0.665)(250,0.665)
    };
    \legend{SimNet, Cosine}
 
\end{axis}
\node[above,font=\small] at (current bounding box.north) {\textsc{Oxford5k}};
\end{tikzpicture} &
\begin{tikzpicture}[scale=0.7]
\large
\begin{axis}[
    xlabel={num. samples},
    xmin=0, 
    xmax=750,
    ymin=0.450, 
    ymax=0.8,
    xtick={0, 250, 500, 750},
    ytick={0.45, 0.55, 0.65, 0.75, 0.8},
    legend pos=north west,
    ymajorgrids=true,
    grid style=dashed,
    legend style={draw=none, fill=none},
]
 
\addplot[
    color={rgb:red,1;green,2;blue,5},
    line width=0.5mm
    ]
    coordinates {
    (0,0.460)(10, 0.457)(50, 0.488)(100, 0.561)(250,0.654)(500,0.709)(750, 0.728)
    };
    
\addplot[
    color=red,
    dashed,
    line width=0.5mm
    ]
    coordinates {
    (0,0.638)(750,0.638)
    };
    \legend{SimNet, Cosine}
 
\end{axis}
\node[above,font=\small] at (current bounding box.north) {\textsc{Paris6k}};
\end{tikzpicture} \\
\end{tabular}
\caption{Domain Adaptation. mAP when using different number of target samples in the training set.}
\label{fig:numSamples}
\end{figure*}
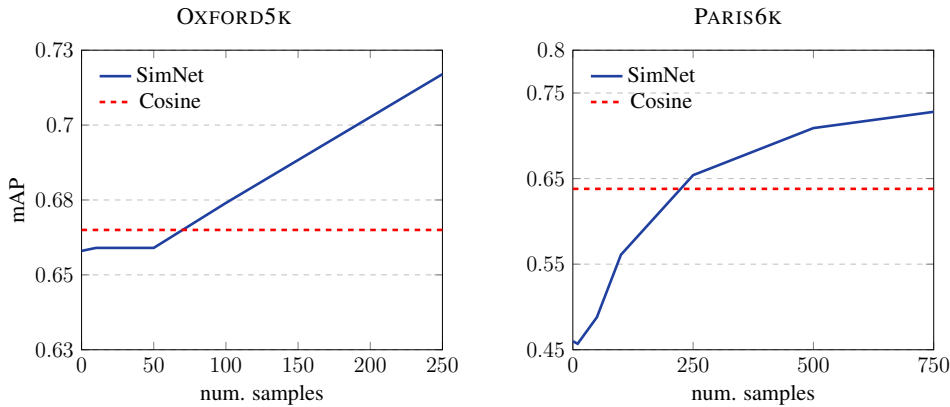

\paragraph{} Results are summarize in Table \ref{tab:tablefull}. Trained similarity networks (SimNet, SimNet*) outperform trained linear methods (OASIS, Linear) in all but one testing datasets. As all Linear, SimNet and SimNet* are trained using the same supervised learning protocol and images, the results suggest that the improvement obtained in our method is not because of the supervision but because of the non-metric nature of the model, which is supposed to fit the human visual perception more accurately. Moreover, when using \textsc{Landmarks-extra} as training dataset, results are boosted with respect to the standard metric, achieving improvements ranging from 20\% (\textsc{Oxford5k}) to 40\% (\textsc{Pairs6k}). When using \textsc{Landmarks-extra500} dataset, our similarity networks also improve mAP with respect to the cosine similarity in the three testing datasets. This indicates that visual similarity can be learnt even when using a reduced subset of the target image domain. However, visual similarity does not transfer well across domains when no images of the target domain are used during training, which is a well-known problem in metric learning systems \cite{kulis2013metric}. In that case, cosine similarity is the best option over all the methods.

\subsection{Image Representation Discussion}
Next, we study the generalisation of our similarity networks when used on top of different feature extraction methods: the output of a VGG16 network \cite{simonyan2014very}, the output of a ResNet50 network \cite{he2016deep}, MAC \cite{tolias2016particular}, RMAC \cite{tolias2016particular} and the model from \cite{radenovic2016cnn}. We compare the results of our networks, SimNet and SimNet*, against cosine similarity. Results are provided in Figure \ref{fig:features}. Our similarity networks outperform cosine similarity in all the experiments, improving retrieval results when used on top of any standard feature extraction method. Moreover, performance is boosted when SimNet* is applied, specially in features with poor retrieval performance, such as ResNets.

\subsection{Domain Adaptation}
We further investigate the influence of the training dataset on the similarity score when it is transfered between different domains or collections of images.
As already noted in  Table  \ref{tab:tablefull},  visual  similarity  does  not  transfer well  across  domains,  and  a  subset  of  samples  from  the  target dataset is required during  training  to  learn  a  meaningful  similarity  function.  This  is  mainly  because similarity  estimation  is  a  problem-dependent  task,  as  the  similarity between  a  pair  of  elements  depends  on  the  data  collection. Thus, in Figure \ref{fig:numSamples}, we explore the effect on performance when we use different subsets of samples from the target collection in addition to the \textsc{Landmarks} dataset. To add relevant samples progressively to the training set, we assign a class label to each image in the \textsc{Oxford5k} and \textsc{Paris6k} collections. These datasets do not provide class labels \textit{per se}, so to overcome this issue, we use the file name of each image as its class label.

\begin{figure*}
\centering
\includegraphics[width=\textwidth]{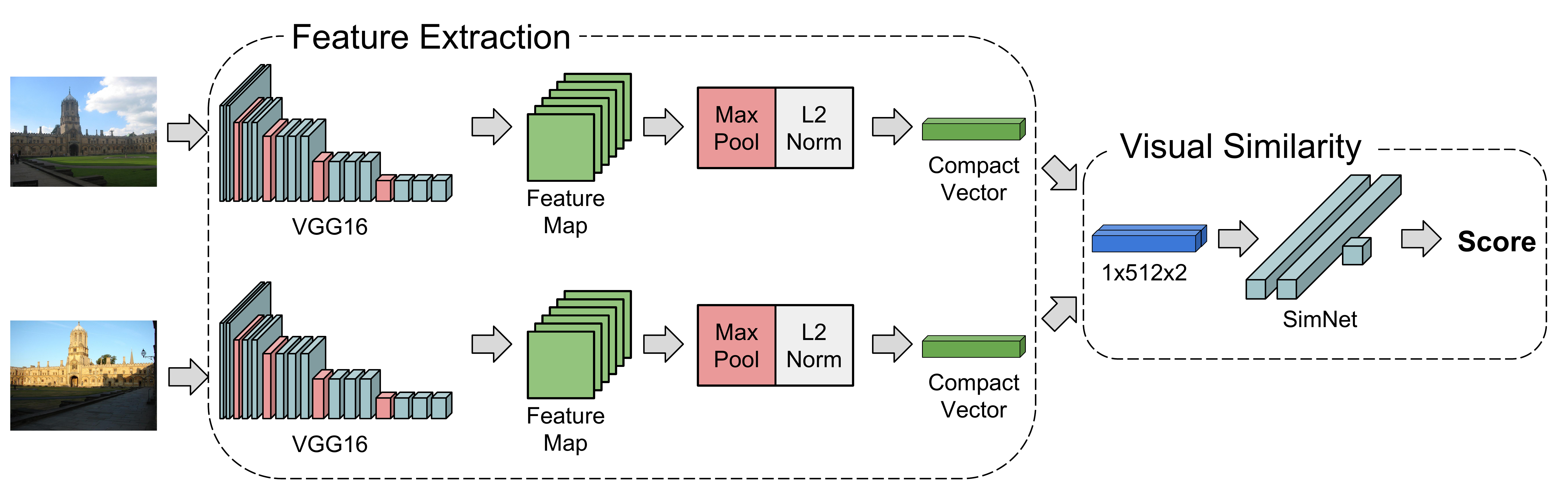}
\caption{End-to-End model. MAC as feature extraction and SimNet as visual similarity.}
\label{fig:end2end}
\end{figure*}

There is a clear correlation between the similarity network performance and the number of samples from the target dataset used during training. Indeed, in agreement with previous work in metric learning \cite{kulis2013metric}, we observe that not considering samples from the target dataset to train a similarity function might be harmful. The similarity network, however, outperforms standard metric results even when a small number of samples from the target collection is used during training: only 100 images from Oxford5k and 250 images from Paris6k are required to outperform cosine similarity in \textsc{Oxford5k} and \textsc{Paris6k} datasets, respectively. This suggests that the similarity network is able to generalise from a small subset of target samples, instead of memorising distances in the training collection.

\subsection{End-to-End Training}

So far, we have isolated the similarity computation part to verify that the improvement in the testing datasets compared to when using other similarity methods is, in fact, due to the visual similarity network. However, with all the modules of the retrieval system being differentiable, an end-to-end training model is also possible. End-to-end methods have been shown to achieve outstanding results in many different problems, including the aggregation of video features \cite{xu2018sequential}, stereo matching \cite{han2015matchnet}, person re-identification \cite{li2014deepreid} or self-driving cars \cite{bojarski2016end}. In image retrieval, however, only end-to-end methods for learning the feature representations are proposed \cite{gordo2017end, radenovic2016cnn}, leaving the final similarity score to be computed with a cosine similarity. In this section we explore a real end-to-end training architecture for image retrieval, which is presented in Figure \ref{fig:end2end}.

\begin{table}
\centering
\vspace{5pt}
\begin{tabular}{ c  c  c  c  c  c  c  c}
   \Xhline{2\arrayrulewidth}
   Features &  Sim. & \textsc{Ox5k} &  \textsc{Pa6k} &  \textsc{La5k}\\ \hline
   MAC & Cosine	 & 0.481 & 0.539 &  0.494 \\
   MAC & \colorbox{lightgray}{SimNet} & 0.509 & 0.683 & 0.589 \\
   \colorbox{lightgray}{MAC} & \colorbox{lightgray}{SimNet} & 0.555 & 0.710 & 0.685 \\ 
   \Xhline{2\arrayrulewidth}
\end{tabular}
\vspace{5pt}
\caption{End-to-End Training. mAP when fine-tunning different parts of the pipeline. In \colorbox{lightgray}{gray}, the modules that are fine-tunned in every experiment.}
\label{tab:resultsEnd2End}
\end{table}

For the feature extraction part, we adopt MAC \cite{tolias2016particular}, although any differentiable image representation method can be used. To obtain MAC vectors, images are fed into a VGG16 network \cite{simonyan2014very}. The output of the last convolutional layer is max-pooled and l2-normalized. For the visual similarity part, we use SimNet with $\Delta = 0.2$. As the whole architecture is end-to-end differentiable, the weights are fine-tunned through backpropagation. We first train the similarity network by freezing the VGG16 weights. Then, we unfreeze all the layers and fine-tune the model one last time. As all the layers have been already pre-trained, the final end-to-end fine-tunning is performed in only about 200,000 pairs of images from \textsc{Landarmarks-extra} dataset for just 5,000 iterations. 

Results are provided in Table \ref{tab:resultsEnd2End}. There is a significant improvement when using the similarity network instead of the cosine similarity, as already seen in the previous section. When the architecture is trained end-to-end results are improved even more, since fine-tuning the entire architecture allows a better fit to a particular dataset. 

\begin{table*}[h]
\centering
\begin{tabular}{ l  c c  c  c  c  c }
\Xhline{2\arrayrulewidth}
     	Method & Dim	& Similarity & \textsc{Ox5k} & \textsc{Ox105k} & \textsc{Pa6k} & \textsc{Pa106k} \\ \hline
 
    Babenko et al. \cite{babenko2014neural} & 512 & L2 & 0.435 & 0.392 & - & - \\
    Razavian et al. \cite{razavian2014cnn} & 4096 & Averaged L2 & 0.322 & - & 0.495 & - \\
    Wan et al. \cite{wan2014deep} & 4096 & OASIS & 0.466 & - & 0.867 & - \\   
    Babenko et al. \cite{babenko2015aggregating} & 256 & Cosine & 0.657 & 0.642 & - & - \\
   Yue et al. \cite{yue2015exploiting} & 128 & L2 & 0.593 & - & 0.59 & - \\
   Kalantidis et al. \cite{kalantidis2016cross} & 512 & L2 & 0.708 & 0.653 & 0.797 & 0.722 \\
   Mohedano et al. \cite{mohedano2016bags} & 25k & Cosine & 0.739 & 0.593 & 0.82 & 0.648 \\
   Salvador et al. \cite{salvador2016faster} & 512 & Cosine & 0.588 & - & 0.656 & - \\
   Tolias et al. \cite{tolias2016particular} & 512 & Cosine & 0.669 & 0.616 & 0.83 & 0.757 \\ 
   Jimenez et al. \cite{jimenez2017class} & 512 & Cosine & 0.712 & 0.672 & 0.805 & 0.733\\
   Ours ($\Delta = 0.8$) & 512 & \textbf{SimNet*} & \textbf{0.808} & \textbf{0.772} & \textbf{0.891} & \textbf{0.818} \\ \hline
   
   \Xhline{2\arrayrulewidth}
\end{tabular}
\vspace{6pt}
\caption{State of the Art Comparison (Off-the-shelf). Dim corresponds to the dimensionality of the feature representation and Similarity is the similarity function.}
\label{tab:sota1}
\end{table*}

\begin{table*}[h]
\centering
\begin{tabular}{ l  c c  c  c  c  c }
\Xhline{2\arrayrulewidth}
     	Method & Dim	& Similarity & \textsc{Ox5k} & \textsc{Ox105k} & \textsc{Pa6k} & \textsc{Pa106k} \\ \hline

   Babenko et al. \cite{babenko2014neural} & 512 & L2 & 0.557 & 0.522 & - & - \\
   Gordo et al. \cite{gordo2016deep} & 512 & Cosine & 0.831 & 0.786 & 0.871 & 0.797 \\
   Wan et al. \cite{wan2014deep} & 4096 & OASIS & 0.783 & - & \textbf{0.947} & - \\ 
   Radenovic et al. \cite{radenovic2016cnn} & 512 & Cosine & 0.77 & 0.692 & 0.838 & 0.764 \\
   Salvador et al. \cite{salvador2016faster} & 512 & Cosine & 0.71 & - & 0.798 & - \\
   Gordo et al. \cite{gordo2017end} & 2048 & Cosine & 0.861 & \textbf{0.828} & 0.945 & \textbf{0.906} \\
   Ours ($\Delta = 0.8$) & 512 & \textbf{SimNet*} & \textbf{0.882} & 0.821 & 0.882 & 0.829 \\
   \Xhline{2\arrayrulewidth}
\end{tabular}
\vspace{6pt}
\caption{State of the Art Comparison (Fine-tunned). Dim corresponds to the dimensionality of the feature representation and Similarity is the similarity function.}
\label{tab:sota2}
\end{table*}

\subsection{State of the Art Comparison}
Finally, we compare our method against several state-of-the-art techniques. As standard practice, works are split into two groups: off-the-shelf and fine-tunned. Off-the-shelf are techniques that extract image representations by using pre-trained CNNs, whereas fine-tunned methods retrain the network parameters with a relevant dataset to compute more accurate visual representation. For a fair comparison, we only consider methods that represent each image with a single visual vector, without query expansion or image re-ranking. Off-the-shelf results are shown in Table \ref{tab:sota1} and fine-tunned results are presented in Table \ref{tab:sota2}. When using off-the-shelf RMAC features, our SimNet* approach outperforms previous methods in every dataset. To compare against fine-tunned methods, we compute RMAC vectors using the fine-tunned version of VGG16 proposed in \cite{radenovic2016cnn}. Accuracy is boosted when our similarity network is used instead of the analogous cosine similarity method \cite{radenovic2016cnn}. SimNet* achieves the best mAP precision in \textsc{Oxford5k} dataset and comes second in \textsc{Oxford105k} and \textsc{Paris106k} after \cite{gordo2017end}, which uses the more complex and higher-dimensional ResNet network \cite{he2016deep} for image representation.

\section{Conclusions}
\label{sec:conclusions}
We have presented a method for learning visual similarity directly from visual data. Instead of using a metric distance function, we propose to train a neural network model to learn a similarity score between a pair of visual representations. Our method is able to capture visual similarity better than other techniques, mostly because of its non-metric nature. As all the layers in the similarity network are differentiable, we also propose an end-to-end trainable architecture for image retrieval. Experiments on standard collections show that results are considerably improved when a similarity network is used. Finally, our work can push performance in image retrieval systems on top of high-quality image features, while it can still be applied with query expansion or image re-ranking methods.

{\small
\bibliographystyle{ieee}
\bibliography{bibliography.bib}
}

\end{document}